# A roadmap to fair and trustworthy prediction model validation in healthcare

Yilin Ning,[1] Victor Volovici,[2,3] Marcus Eng Hock Ong,[4,5,6] Benjamin Alan Goldstein,[4,7] Nan Liu[1,4,8]*

[1]Centre for Quantitative Medicine, Duke-NUS Medical School, Singapore 169857, Singapore
[2]Department of Neurosurgery, Erasmus MC University Medical Center, 3015 GD Rotterdam, The Netherlands
[3]Department of Public Health, Erasmus MC, 3015 GD Rotterdam, The Netherlands
[4]Programme in Health Services and Systems Research, Duke-NUS Medical School, Singapore 169857, Singapore
[5]Health Services Research Centre, Singapore Health Services, Singapore 169856, Singapore
[6]Department of Emergency Medicine, Singapore General Hospital, Singapore 169608, Singapore
[7]Department of Biostatistics and Bioinformatics, Duke University, Durham, NC 27710, USA
[8]Institute of Data Science, National University of Singapore, Singapore 117602, Singapore

**\*Correspondence to:**
Nan Liu
Centre for Quantitative Medicine, Duke-NUS Medical School, 8 College Road, Singapore 169857, Singapore
Email: liu.nan@duke-nus.edu.sg
Phone: +65 6601 6503

## Summary

A prediction model is most useful if it generalizes beyond the development data with external validations, but to what extent should it generalize remains unclear. In practice, prediction models are externally validated using data from very different settings, including populations from other health systems or countries, with predictably poor results. This may not be a fair reflection of the performance of the model which was designed for a specific target population or setting, and may be stretching the expected model generalizability. To address this, we suggest to externally validate a model using new data from the target population to ensure clear implications of validation performance on model reliability, whereas model generalizability to broader settings should be carefully investigated during model development instead of explored post-hoc. Based on this perspective, we propose a roadmap that facilitates the development and application of reliable, fair, and trustworthy artificial intelligence prediction models.

# Introduction

With the increasing availability of electronic health-related databases, computational resources and advanced modeling techniques such as artificial intelligence (AI), there is growing interest to develop data-driven prediction models to aid real-life decision making, with the aim to capture useful patterns in the data that are not obvious to humans.[1] An important follow-up step is to evaluate prediction models using new data not involved in model development, to ensure that the patterns learnt are not restrictive to the development data, but can generate useful predictions for a wider population. Such model evaluation, generally referred to as "external validation", has therefore been a main focus in model development and reporting guidelines (e.g., TRIPOD[2] and its AI extension,[3] STARD-AI,[4] and DECIDE-AI[5]).

Despite the importance of external validation for developing practically useful prediction models, there lack clear guidelines on the choice of evaluation datasets, which in turn affects the expectation and interpretation of validation results. For example, when a prediction model is developed using samples from *population A*, good performance *is expected* when externally validated on new data from *population A* and *may be observed* for data from *population B* with similar characteristics, but it becomes unclear when evaluated using data from *population C* that might have notable differences in population characteristics and possibly misalignments in predictors available. Figure 1 visually summarizes the three external validation settings and their different implications.

**Figure 1.** Visual illustration of three types of external validation for a prediction model intended for local use in *population A*. *External validation I* that uses new data from *population A* is appropriate for assessing model reliability. *External validation II* using external data from *population B* with similar characteristics may be reasonable but not necessarily useful. *External validation III* using external data from *population C* with very different characteristics has questionable implications and must be interpreted with caution.

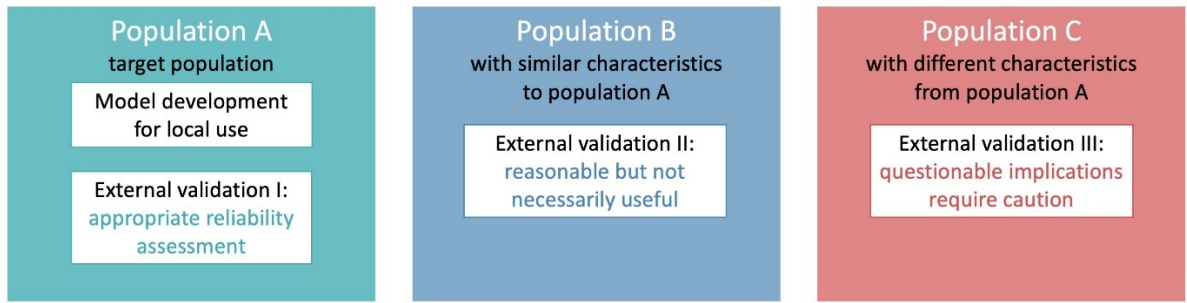

When discussing the performance of a prediction model, evidence is often aggregated across validation studies without accounting for differences in application settings and sample characteristics, which may result in biased assessment of model reliability. A recent work highlighted that if a model is intended for local application within *population A*, validations using external data from *populations B and C* may not be useful,[6] but is it useful to extend the model to accommodate *populations B and C* with diverse characteristics? In this paper we distinguish between types of external validations by practical implications, and propose a roadmap for external validation that focuses on fair and trustworthy healthcare applications.

**Three types of external validation**

External validation of prediction models assesses their *generalizability* (Box 1) beyond the development data, and the choice of validation data reflects the extent of generalizability expected. A prediction model is usually developed for a specific outcome and *target population* (i.e., the type of individuals the model is *intended* to apply to), and is ideally developed using a *representative sample* (i.e., a group of individuals that fairly represent the characteristics of the target population). In the hypothetical example in Figure 1, researchers were interested in developing a prediction model for the initial resuscitation outcome (i.e., return of spontaneous circulation [ROSC]) after out-of-hospital cardiac arrest in a country, and the model was developed using data from the national registry that represented multiple emergency medical service (EMS) systems across the country. In this case, the target population is all cardiac arrest cases in the country (*population A*), and the registry data used for model development is a (hopefully representative) sample from the target population.

**Box 1.** Glossary of model development and validation.

| |
|---|
| *Model development*: the process to develop a prediction model from a dataset (i.e., *development data*), generally divided into a training step that uses a subset of samples in the development data to quantify model parameters, and an *internal validation* step that uses the rest of samples to assess model performance. |
| *Target population*: the collection of all subjects a prediction model is intended to apply to. Development data is a sample from the target population, preferably a representative sample that fairly represents the characteristics of the whole population. |
| *Generalizability*: the ability to generate useful predictions for subjects not included in the development data. |

> *External validation*: evaluation of an existing prediction model using new samples not involved in model development. A key approach to assess generalizability beyond development data to the target population.

When discussing different types of external validation of prediction models, researchers often distinguish between geological external validation using data from a different geographic location, and temporal external validation using data from the same location but a different time period, among others.[7,8] In this section, we identify and discuss three general types of external validation based on a different dimension, i.e., the relationship between validation data and the target population.

**External validation using local data from target population**

Intuitively, a useful model should work well for the outcome and population it was developed for. External validation using new data from the target population directly assesses this, where the validation data should be a representative sample of the target population,[9,10] with the same set of variables and same outcome definition as the development data.[9] *External validation I* in Figure 1 provides such an example, where the prediction model was evaluated using a different sample from *population A* that has not been exposed to the prediction model, e.g., new data from the national registry in the following years (temporal external validation), or data from EMS systems not yet in the national registry (geological external validation), which are consistent with the recommendation of using data that is different yet "plausibly related".[7] As highlighted in the current literature, such external validation is necessary to ensure that a model is generalizable beyond the development data and is reliable for its intended applications (unless when the development data is sufficiently representative),[6] and unsatisfactory model performance should be addressed before its real-life applications.

**External validation using external data from different populations**

While external validation using local data has straightforward implications, in practice researchers often find it necessary (or even preferable) to evaluate a prediction model using external data not from the target population, e.g., by using data from a different region or even a different country. Generally, characteristics of such external data differ from the target population to varying extents, and Figure 1 illustrates two typical examples. External data in *External validation II* (Figure 1) has similar characteristics as the target population, e.g., cardiac arrests in a neighboring country with similar demographics and EMS systems, therefore it may be reasonable to expect (or demand) good predictive performance in the evaluation. In *External validation III*, however, the data source has known differences in characteristics from the target population (e.g., cardiac arrests in a different country with known differences in demographics, EMS systems and medical practice), making it unclear whether good performance *can (or should)* be expected. One reason is that unless specifically designed, prediction models cannot establish what have caused the outcome,[11] but instead leverage on patterns most frequently co-occur with the outcome in the development data, which are often confounded by other (unmeasured) factors and do not necessarily translate to other populations.

**Limitations of validation using external data**

Although we separately discussed the implications for *External validation II and III* due to different extents of data differences from target population, such a distinction is not easily made in real-life applications. To illustrate this, we look at external validations of the ROSC after cardiac arrest (RACA) score, which was developed in 2011 from the German Resuscitation Registry[12] and worked well for EMS system evaluation and comparison in the country.[13]

A recent work validated the RACA score in a Pan-Asian population with a considerably lower ROSC rate (8.2%, compared to 43.8% in Germany) and notable differences in several important predictors (e.g., proportion of shockable initial rhythm, witnessed cardiac arrest and bystander cardiopulmonary resuscitation [CPR]).[14] Consequently, the constant term in the RACA score, which reflects the baseline ROSC rate in the population, needed to be updated to better predict the probability of ROSC for the Pan-Asian population.[14] Compared to the Pan-Asian study, European validation studies[15–17] differed less from the original RACA study in population characteristics, yet differences were still observed in ROSC rates (ranging from 21% to 50%) and characteristics of a few predictors.[17] Intuitively, smaller differences in the ROSC rate may imply more similar settings to the RACA study and therefore better validation results (e.g., better calibration). Interestingly, two validation studies with ROSC rates of 44% and 50% reported suboptimal calibration for patient subgroups,[15,17] yet a validation study with 28% ROSC reported good calibration.[16] Hence, finding external data that is "similar enough" to the target population is non-trivial, and data differences can have unanticipated impact on model performance, making the implications unclear.

**Roadmap for fair and trustworthy external validation**

Discrepancies in sample characteristics between external data and target population can have arbitrary impact on model performance and result in mixed findings.[6,10] However, some works consider evaluation using external data to be stringent and highly encouraged due to the difference in population characteristics in evaluation and development settings.[18,19] We propose an alternative roadmap for fair and trustworthy external validation using local data from the target population (see Figure 2), which allows detailed evaluation of model performance and provides direct evidence to guide model improvement and suitable real-life implementation.

**Figure 2.** Roadmap for fair and trustworthy external validation.

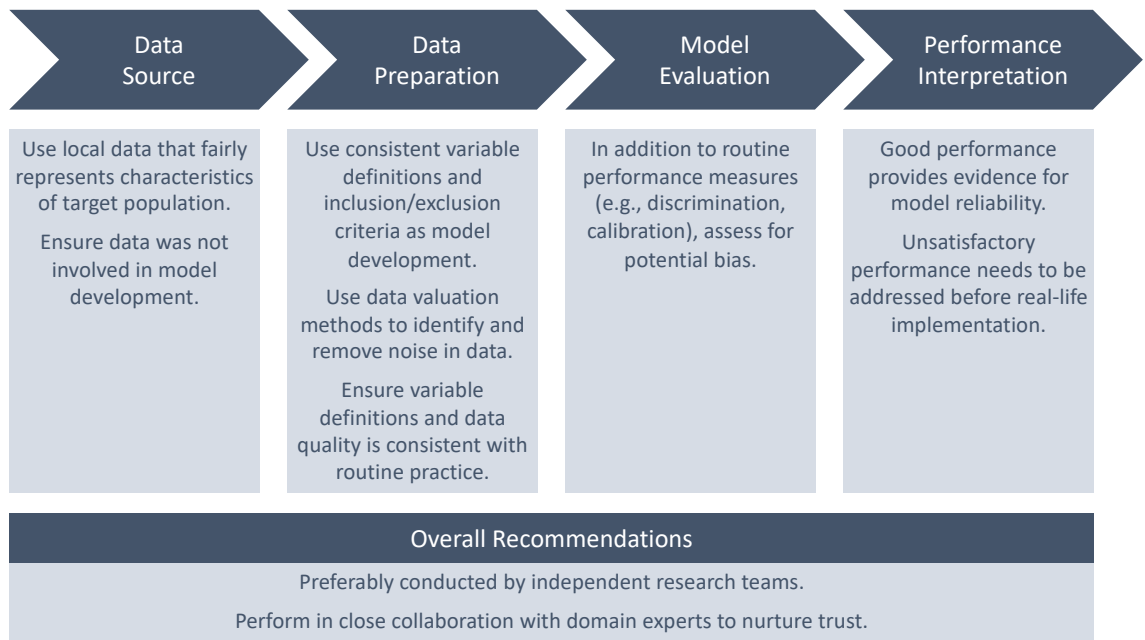

**Assess model reliability via external validation using local data**

As required in the TRIPOD reporting guideline,[2] researchers should explicitly define the target population when reporting prediction models. When the target population comprises multiple centers (e.g., all EMS systems in a country), the heterogeneity across centers should be investigated and accounted for during model development instead of assessed and addressed post-hoc in external validations.[8] Subsequent external validations should evaluate model reliability for the intended purposes (which have been proven feasible during model development) by using new data that is representative of the target population and was not involved in model development. As recently highlighted, using representative samples in model evaluation is crucial for ensuring model validity and accurate risk prediction,[10] and it is also an important consideration when inspecting model fairness and potential bias.[20]

Definition of outcome and predictors in the validation data and the inclusion/exclusion criteria applied should be consistent with those adopted during model development. PROBAST (Prediction model Risk Of Bias ASsessment Tool)[21] provides a checklist to help researchers identify (and hence reduce) potential bias in validation data. In addition to routine data quality inspections, researchers may use recently developed data valuation methods to identify harmful observations and reduce noise.[22] However, the general data quality and variable definitions should also be consistent with routine practice to be practically useful.[10]

As recommended, external validation is preferably performed by independent research teams to better assess replicability,[23] and model inadequacies identified should be addressed via model adjustment or re-development, e.g., using methods previously described.[18] In addition to routine assessments of predictive performance (e.g., discrimination, calibration and decision-curve analysis),[7] researchers may refer to recent methodological developments[20] to investigate for potential bias and unfairness in predictions. Extensive evaluation in the local context in close collaboration with domain experts also help resolve concerns and nurture trust in the prediction model for real-life implementation.

**Perform external validation sparingly when local data is unavailable**
Although we recommend external validation using representative samples from the target population, in practice it may be difficult to curate validation data with sufficient sample size due to limited resources. In such cases, researchers should only consider external validation when there are compelling practical needs, e.g., to retrospectively label the application populations of existing machine learning-based medical devices.[24] Researchers may consider external data that is known to have similar characteristics to the target population to minimize data impact on model performance and clearly describe any data differences,[6] yet validation using data from the target population is still needed for formal model assessment. When a model is evaluated using data with known differences from the target population, findings must be interpreted with caution: poor model performance may result from data differences instead of model inadequacy, whereas good performance can be chance findings instead of evidence for generalizability.

**Develop new model for application to external data**

An important aim of external validation is to assess if an existing model can be applied to a different population, possibly after a series of complex model updates.[6,18] However, time and resources may be better spent by developing new models for the new population. In addition to genuine differences in population characteristics, differences in data recording systems and medical practice can lead to misalignments in detailed information that affects the performance of external models. In the RACA evaluation example, the Pan-Asian and German data defined different categories for location of cardiac arrest that could not be exactly mapped.[14] In a subsequent study, the RACA score was outperformed by a local score developed using Pan-Asian data that accounted for the differences in variable definitions, information resolution, and medical practice.[25]

A rationale for re-purposing existing models (possibly after adjustments) is to build upon existing evidence in the literature.[18] However, instead of directly applying or adapting an existing model, we propose to build upon the data preparation and model development process established by existing high-quality studies, and develop new models using local data that are tailored to local concerns and needs. Recent developments in common data models[26] facilitate standardisation of data preparation pipelines, and automated machine learning frameworks improve reproducibility of model development steps for easy-to-interpret scoring systems[27–29] and flexible black-box models.[30]

## Conclusions

As an important tool for assessing model reliability and validity, external validation must be performed and interpreted with caution. We highlight limitations of external validation using external data beyond the target population of the prediction model, and proposed a roadmap for external validation that focuses on model reliability, fairness and trustworthiness. External validations should assess model reliability and fairness for intended applications using new data from a clearly defined target population, and in close collaboration with domain experts to nurture trust. Model performance in different populations should be interpreted with caution, as it is arbitrarily affected by the differences in population characteristics. Hence, instead of seeking to expand the application of an existing model post-hoc via external validation, we advise researchers to replicate the data processing and model development pipelines of well-established models to train models tailored to new needs.

**Contributors**

NL conceived the idea for the Viewpoint. YN developed the first draft. All authors edited the manuscript, read and approved the final manuscript, and had final responsibility for the decision to submit for publication.

**Declaration of interests**

The authors declare no conflicts of interest.

**Acknowledgements**

This work was supported by the Duke-NUS Signature Research Programme funded by the Ministry of Health, Singapore. YN is supported by the Khoo Postdoctoral Fellowship Award (project no. Duke-NUS- KPFA/2021/0051) from the Estate of Tan Sri Khoo Teck Puat. The funders were not involved in the study design, collection, analysis, and interpretation of data, nor did they have a role in the writing of the paper and decision to submit the paper for publication. The authors have not been paid to write this article by a pharmaceutical company or other agency. All authors had access to the data in the study and had final responsibility for the decision to submit for publication.